\documentclass[journal]{IEEEtran}
\ifCLASSINFOpdf
\else
\fi
\usepackage{times}
\usepackage{epsfig}
\usepackage{graphicx}
\usepackage{amsmath}
\usepackage{amssymb}
\usepackage{tabulary}
\usepackage{multirow}
\usepackage{booktabs}       
\usepackage{float}
\usepackage{stfloats}
\usepackage{bm}
\usepackage{lineno}

\begin{document}
%
\title{Residual Pyramid Learning for Single-Shot Semantic Segmentation}
%
%
%

\author{Xiaoyu~Chen,
        Xiaotian~Lou,
        Lianfa~Bai,
        and~Jing~Han
\thanks{The authors are with the school
of Electronic and Optical Engineering, Nanjing University of Science and Technology, Nanjing,
Jiangsu, 210091 China e-mail: 115104000466@njust.edu.cn; 18069984123@njust.edu.cn; blf@njust.edu.cn; eohj@njust.edu.cn}}

%
%

\markboth{Journal of \LaTeX\ Class Files,~Vol.~14, No.~8, August~2015}%
{Shell \MakeLowercase{\textit{et al.}}: Bare Demo of IEEEtran.cls for IEEE Journals}
%



\maketitle

\begin{abstract}
   Pixel-level semantic segmentation is a challenging task with a huge amount of computation, especially if the size of input is large. In the segmentation model, apart from the feature extraction, the extra decoder structure is often employed to recover spatial information.  In this paper, we put forward a method for single-shot segmentation in a feature residual pyramid network (RPNet), which learns the main and residuals of segmentations by decomposing the label at different levels of residual blocks. Specifically speaking, we use the residual features to learn the edges and details, and the identity features to learn the main part of targets. At testing time, the predicted residuals are used to enhance the details of the top-level prediction. Residual learning blocks split the network into several shallow sub-networks which facilitates the training of the RPNet.  We then evaluate the proposed method and compare it with recent state-of-the-art methods on CamVid and Cityscapes. The proposed single-shot segmentation based on RPNet achieves impressive results with high efficiency on pixel-level segmentation.
\end{abstract}

%
\IEEEpeerreviewmaketitle

\section{Introduction}
%
%
%
%
\IEEEPARstart{A}{utonomous} driving technology has been developing rapidly to increase the transportation efficiency and improve driving experience. Scene parsing from images in autonomous driving systems is getting more and more attention as cameras are the most accessible and inexpensive sensors. With recent advances of deep neural networks (DNNs)~\cite{lecun1998gradient}, the performance of semantic segmentation has been increased greatly. However, how to balance the performance and new huge computation cost, brought by deep neural networks, has became a new problem in real-time systems.

Neural network is often constructed in the shape of pyramid to increase displacement invariability and reduce computation. Recently, researchers use the hidden features of pyramid networks to produce more powerful feature for more complex tasks such as super resolution and semantic segmentation tasks, as spatial details is on the decrease from the bottom to the top of networks~\cite{7803544, Lin:2017:RefineNet, Yang_2018_CVPR}.

Semantic segmentation is to recognize every pixel of inputs. Therefore it is essential to construct feature maps including features of all pixels. The popular encoder-decoder models use a pyramid network to train an encoder to approximate low-resolution target and employ an inverted pyramid network followed by encoder to reconstruct the high-resolution target, where unpooling~\cite{Zeiler2014VisualizingAU} and deconvolution ~\cite{7410535} are often used in decoders. Because the details are missing during encoding process, the effect to recover details is limited and much extra computation is entailed in decoders. To achieve a balance between efficiency and reliability, some methods delete decoder structure and interpolate the low-resolution result directly at the expense of details ~\cite{Yu2018BiSeNetBS, 7913730}.

Different levels of features in neural networks represent different levels of semantics. High-level features have strong semantics while low-level features have rich spatial details. Therefore the feature fusion of features from different levels is often adopted in decoder structure.

Although some methods are proposed to uses low-level features for prediction as supplements, simply integrating the low-level and high-level feature will accumulate information redundancy~\cite{10.1007/978-3-319-24574-4_28,Ghiasi2016LaplacianRA}. Since the different levels of features are mutually complemental, we attempt to use the features to approximate the target separately and combine the outputs of them to get the full target.

ResNet~\cite{He2016DeepRL} is a excellent example of feature separation, which is proposed to ease the training of deep networks by adding identity mapping which separates the feature into residual part and identity part in one block:
\begin{gather}\label{eq:transform}
\bm{identity}_{l + 1} = \bm{identity}_{l} + \bm{res}_{l}\nonumber \\
 \sim \,\bm{identity}_{l} = \bm{identity}_{l + 1} + (-\bm{res}_{l}),\label{eq:f}
\end{gather}
where $\bm{identity}_{l+1}$ and ($-\bm{res}_{l}$) can be seen as independent part of $\bm{identity}_{l}$. Different from the features pyramid in the plain networks, the features of ResNet form a feature residual pyramid~\cite{He2016IdentityMI}, which is similar to the process of Laplacian residual pyramid~\cite{Burt1987The}:
\begin{equation}
\bm{p}_{l} = \bm{p}_{l+1} + \bm{pres}_{l}\,\label{eq:p}
\end{equation}
where $\bm{identity}_{l + 1}$ and $\bm{p}_{l + 1}$ respectively represent the identity features in ResBlocks and the laplacian pyramid images with higher resolution at $(l+1)$-th level, $\bm{identity}_{l}$ and $\bm{p}_{l}$ respectively represent that with lower resolution, $\bm{res}_{l}$ and $\bm{pres}_{l}$ represent the residuals. For simplicity, upsampling is omitted before pixel-wise sum in Eqn.(\ref{eq:f}) and Eqn.(\ref{eq:p}).

In general, the low-level features in the network with small receptive field contain the information of local texture, which focus on the details and edges of instances, while the high-level features with big receptive field focus on the overall attribute of bigger blobs~\cite{Zeiler2014VisualizingAU}. In ResNet-like networks, hidden features of ResBlocks are further designed as residual feature ($\bm{res}_{l}$) that represents residual information.

From Eqn.(\ref{eq:f}) we can see that ($-\bm{res}_{l}$) is lost in feed-forward pass along with the reduction of spatial details. With the characteristics of features in different ResBlocks in the network, We can just use ($-\bm{res}_{l}$) to retrieve the lost details ($\bm{pres}_{l}$).

For this purpose, we decompose the target of segmentation to train the residual pyramid network at training time, in which the details and main of the target are separated and learned respectively with the residual features and high-level identity features. Then at testing time, the approximated target residual pyramid by these features can be reconstructed to full target.

In the overall forward pass of the network, the input is separated into multiple independent residual information branches and main low frequency branch, and then we collect these branches and use them to approximate different parts of the target. By shortening the paths of gradient propagation in residual branches, the residuals learning blocks also facilitate the training process. Besides, the input only passes through single pyramid network without decoding process, which reduce much computation and increase the efficiency significantly.

The concept of single-shot was firstly proposed in object detection task to describe the detectors with single neural network~\cite{liu2016ssd,Redmon_2016_CVPR}. We quote this concept for our single residual pyramid network, and distinguish it from encoder-decoder methods. In object detection task, the single-shot detector often has higher efficiency than models with multiple networks. The single-shot detector SSD, one of the fastest detectors, uses a single backbone network to obtain different levels of features and generate different scales of boxes on these features for multi-scale detection, saving much computation and reducing latency. In order to enhance the efficiency of segmentation, we use this idea to approximate different levels of target residuals at different layers of backbone network. The single-shot RPNet only entails several simple convolution layers apart from the original pyramid network.

To improve the efficiency of perceiving road environment in intelligent vehicles, this paper puts forward a residual pyramid learning network (RPNet) to learn the residual pyramid of target and implement the single-shot segmentation. We design a loss function to train residual features to residual target. Instead of decomposing labels directly, the key of training a PRNet is to reconstruct the full predicted targets at different levels to compute loss with full labels of different scales in residual pyramid. We implement the RPNet on different backbone networks to evaluate the proposed method.

This paper makes the following contributions:

\textit{(i)}We propose a novel single-shot structure for segmentation to predict different parts of target in a single pyramid feed-forward pass which improves efficiency significantly.

\textit{(ii)}We use residual features to predict multi-level residuals of target by designing a residual loss function and conduct top-down level-wise training. The performance on small objects and details has a obvious promotion.

\textit{(iii)}We design variations of residuals and predictor based on RPNet, and expand the application for arbitrary structures beyond the ResNet-like networks.

\textit{(iiii)}We setup RPNet on existing high-speed segmentation networks and achieve both accuracy and efficiency improvements.

\section{Related Work}
Application of semantic segmentation in autonomous driving systems require high accuracy as well as low latency to ensure driving security. However, the computing source is limited on intelligent vehicles, so improving accuracy with limited computation and maintaining accuracy while reducing computation are two important directions.

\subsection{Multi-Path and Single-Shot structure}
As repeated downsampling operateors in pyramid neural networks lead to a significant decrease in image resolution, many methods are proposed to recover the details from low-level feature.

Encoder-decoder structure is often used to recover the spatial details in semantic segmentation. U-type structure, proposed in U-Net~\cite{10.1007/978-3-319-24574-4_28}, use multi-path to help low-level features skip the middle layers and be combined with the refined high-level feature in the decoder, which enhance the performance on details. FC-DenseNet~\cite{8014890} extended DenseNets~\cite{8099726} in U-type structure and improve the upsampling path in decoder to reduce computation.

Besides of U-type structure, multiple feature fusion is also commonly used to recover details. RefineNet~\cite{Lin:2017:RefineNet} proposes a generic multi-path refinement network that fuses multi-resolution features from different layer to generate high-resolution and high-quality results. However, before multi-resolution fusion, many convolutions are added, and the speed on 512x512 images is only 20fps. Light-Weight RefineNet~\cite{nekrasov2018light} increases the speed by modifying the RCU and CRP module in original RefineNet, and operates at the speed of 55fps.

BiSeNet~\cite{Yu2018BiSeNetBS} uses bilateral segmentation network, Spatial Path and Context Path to achieve both rich spatial information and sizeable receptive field. In order to improve efficiency, the BiSeNet uses a decoder of interpolation. ICNet~\cite{zhao2018icnet} sets up a image cascade network with multi-resolution branches under proper label guidance to reduce much computation in pixel-level segmentation inference.

Multi-path fusion strategy often uses the full feature of each layer, so there is much information and computing redundancy.

Decoder of single interpolation is often adopted to avoid computing redundancy such as DeepLabV3~\cite{Chen2017RethinkingAC} and ESPNet~\cite{DBLP:journals/corr/abs-1803-06815}. Here we regard the methods with interpolation decoder as one of single-shot structures. But this single-shot structure can result in the loss of details, and a post processing with conditional random field (CRF)~\cite{krahenbuhl2011efficient} is often used to refine the coarse segmentation, which adds more extra latency.

\subsection{Residual learning for pixel-level approximation}

Residual learning is first proposed in classification network~\cite{He2016DeepRL} to improve the network degradation and gradients vanishing problems. With many advantages in network training, the concept of residual learning has been migrated to other tasks, such as super resolution and semantic segmentation tasks.

Super resolution can be considered as a process of details restoration, so learning sparse residuals is more efficient than learning full images itself of higher resolution. VDSR~\cite{7780551} sets up a deep neural network to learn the high frequency details, which achieve great improvement on accuracy. LapSRN~\cite{8434354} expands the conception to cascade residual learning blocks and progressively reconstructs the sub-band residuals of high-resolution images at multiple pyramid levels and further increase the performance.

\begin{figure*}[hb]
\begin{center}
   \includegraphics[width=0.85\linewidth]{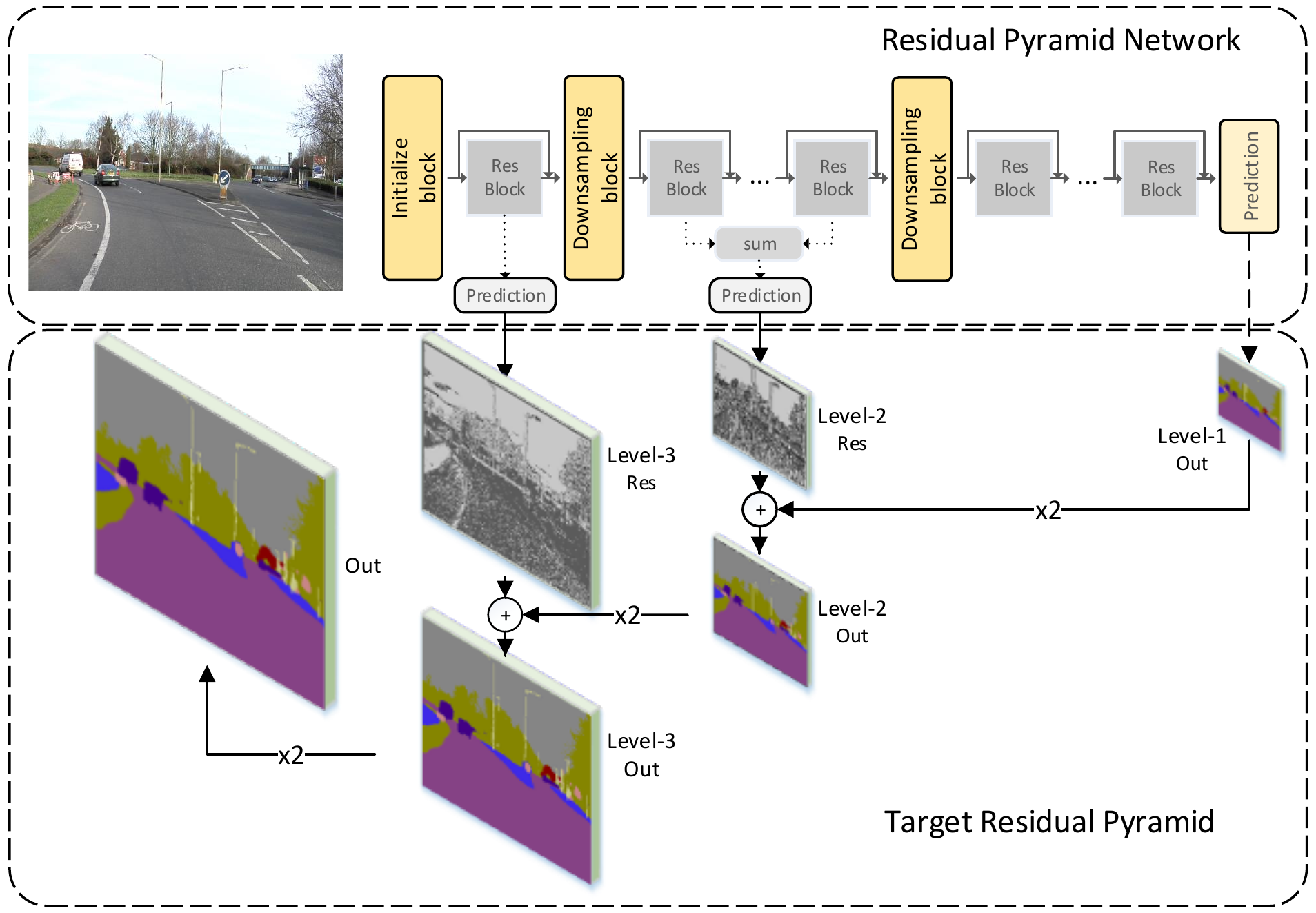}
\end{center}
   \caption{Single-shot architecture based on RPNet.}
\label{fig:arch}
\end{figure*}

In semantic segmentation tasks, there are also many attempts of residual learning. LRR~\cite{Ghiasi2016LaplacianRA} uses low-resolution results to generate a boundary mask for high-resolution feature. Then high-resolution boundary is predicted to refine the result. Global-Local-Refinement~\cite{ijcai2017-479} iteratively predicts global residuals using the input which concatenates the original image and confidential map.

All these methods have proved that residual learning can not only ease the optimization of network, but also improve the efficiency of the approximation.

\subsection{Optimizing networks with multiple loss layers}

Many networks have multiple loss layers for multiple tasks or stronger supervision. Weighted sum of the losses is often used for training in these networks.

PSPNet~\cite{8100143} and BiSeNet~\cite{Chen2017RethinkingAC,Yu2018BiSeNetBS} use mid predictors to construct auxiliary loss layer followed by hidden layer. The auxiliary loss, illustrated in~\cite{8100143}, helps the optimization during learning process, while the master branch loss takes the most responsibility. The ablation study in~\cite{8100143} is also presented to help decide which weight values to use. The Impatient DNNs~\cite{DBLP:journals/corr/AmthorRD16} attempts to use multiple early prediction layers to deal with dynamic time budgets during application, where the intermediate predictors are learned jointly with the weighted losses. In the paper~\cite{DBLP:journals/corr/AmthorRD16}, the authors compare different weights per loss component and choose the best weights to train the network.

In some specific designed network, joint training of multiple losses will break the structure of learned feature. In LRR~\cite{Ghiasi2016LaplacianRA}, the coarse and fine semantic segmentations are predicted from top to down in the network, where fine segmentation predictions depend on the higher coarse predictions. So the level-wise training is necessary in LRR.

\subsection{Improvement for Real-time Segmentation}
Real-time segmentation models prefer thin networks with fewer filters so that computing cost can be reduced such as ENet~\cite{Paszke2016ENetAD}. But simply reducing computation will lead to degradation in performance.

In segmentation task, decoder structure is often removed at the expense of spatial details to reduce computing cost in some methods such as DeepLab v3 and BiSeNet~\cite{Chen2017RethinkingAC,Yu2018BiSeNetBS}.

Another way is to optimize convolution blocks. ERFNet~\cite{Romera2018ERFNetER} uses residual connections and factorized convolutions~\cite{DBLP:journals/corr/WangLF16b} to maintain efficiency and accuracy. DeepLab v3+~\cite{Chen2018EncoderDecoderWA} proposes Atrous Separable Convolution to speed up standard convolution. ESPNet~\cite{DBLP:journals/corr/abs-1803-06815} decomposes a standard convolution into a point-wise convolution and a spatial pyramid of dilated convolutions to improve efficiency. BiSeNet~\cite{Chen2017RethinkingAC,Yu2018BiSeNetBS} uses shallow Spatial Path and Context Paths to generate high-resolution feature and sufficient receptive field, and then combines the two paths to predict the target. This two-path structure can transfer the computation of depth to two sub-networks and improve the parallelism of sub-networks.

Therefore, smart operators and improved structure is needed in real-time segmentation with the consideration of computing complexity and resource utilization.

\section{Residual Pyramid Networks}

The basic architecture of the proposed Residual Pyramid Network (RPNet) is based on a ResNet-like backbone network, which is shown in Fig.~\ref{fig:arch}. More complicated structures will be discussed in following chapters. An example of backbone network of ENet~\cite{Paszke2016ENetAD} encoder is shown in Table.(\ref{tab:Backbone network}), which consists of three downsampling blocks, three residual learning blocks and the main leaning block for segmentation. In RPNet, the features of different ResBlocks in the specific level are added up and passed through a predictor to compute residuals, while the top output of the backbone is used to predict the small scale of segmentation. Finally, the predicted residual pyramid is reconstructed to get full segmentation.

\subsection{Construct a Residual Leaning Block}
In ResNet, stacked ResBlocks can be expressed as:
\begin{gather}
\bm{y}_{l} = h(\bm{x}_{l}) + \mathcal{F}(\bm{x}_{l}, \mathcal{W}_l), \label{eq:resunit1}\\
\bm{x}_{l+1} = f(\bm{y}_{l}) \label{eq:resunit2}.
\end{gather}
where $\bm{x}_{l}$ and $\bm{x}_{l+1}$ are input and output of the $l$-th block, $\mathcal{F}$ is a residual function with the parameters $\mathcal{W}_l$, $h(\bm{x}_{l})$ is an identity mapping and $f$ is an activation. Regardless of the activation $f$, we can recursively put Eqn.(\ref{eq:resunit2}) into Eqn.(\ref{eq:resunit1}) and obtain the output $\bm{x}_{L}$:
\begin{equation}
\bm{x}_{L} = \bm{x}_{l} + \sum_{i=l}^{L-1}\mathcal{F}(\bm{x}_{i}, \mathcal{W}_{i}), \label{eq:additive}
\end{equation}
In Eqn.(\ref{eq:additive}), the residual function $\mathcal{F}$ is the part to be trained and residual feature can be defined as:
\begin{equation}
\bm{res} = \bm{x}_{L} - \bm{x}_{l} = \sum_{i=l}^{L-1}\mathcal{F}(\bm{x}_{i}, \mathcal{W}_{i}), \label{eq:respart}
\end{equation}
Here we get the residual feature of single level and across the network we can get multiple levels of residual features.

\begin{figure*}[hpt]
\begin{center}
   \includegraphics[width=0.80\linewidth]{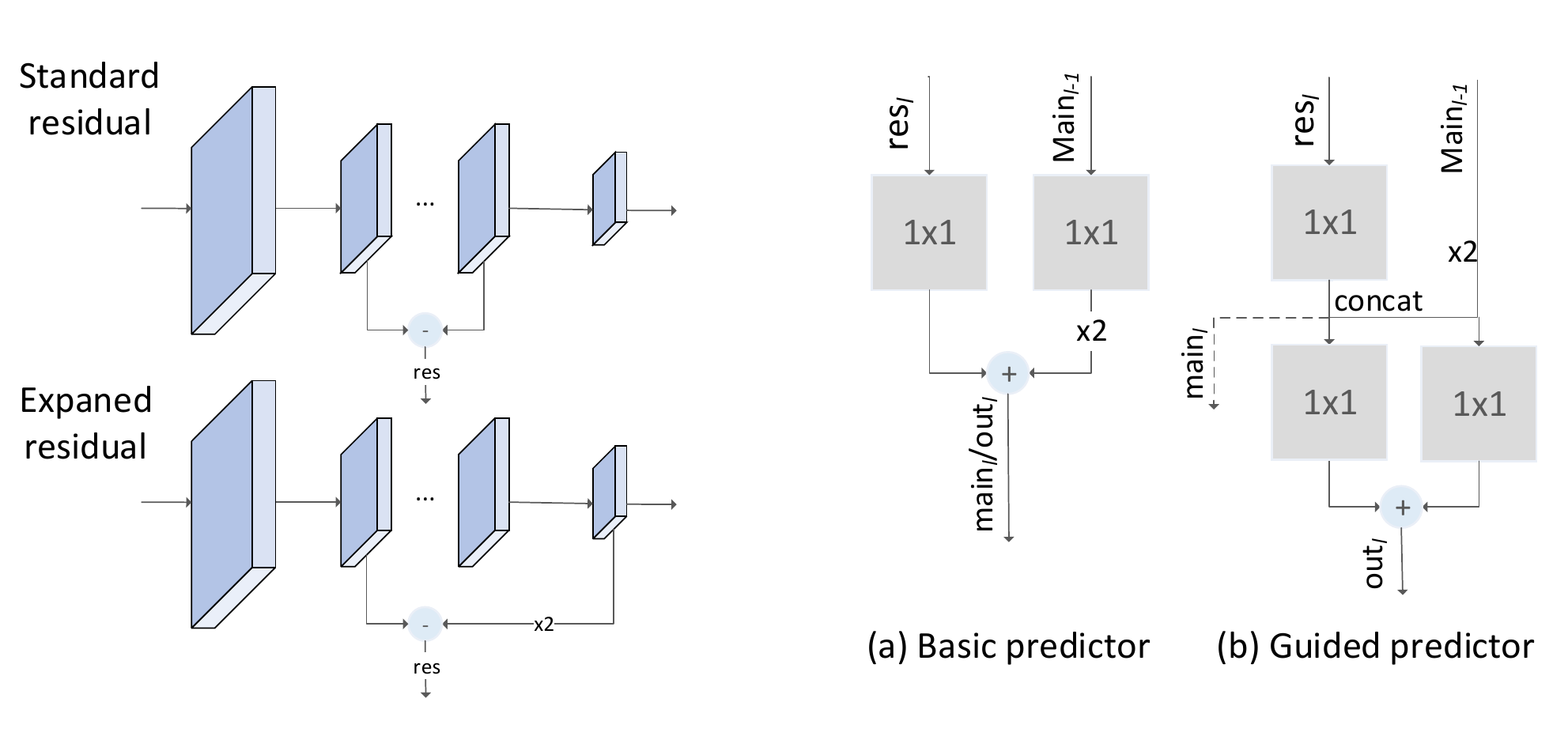}
\end{center}
   \caption{Variations of residuals and predictors. Left-up is standard residuals (\textbf{sr}) and left-down is expanded residuals (\textbf{er}); basic (\textbf{bp}) and guided (\textbf{gp}) predictors are on the right.}
\label{fig:rr}
\end{figure*}

We regard the $(-\bm{res})$ as one stream with high-frequency information of the input, which is to be subtracted from the input feature $\bm{x}_{l}$ to get the other stream of main part $\bm{x}_{L}$, which maintains more low-frequency information:
\begin{equation}
\bm{x}_{l} = \bm{x}_{L} + (- \bm{res}).\nonumber
\end{equation}

Generally, the $\bm{res}$ of residual block shown in Equition.(\ref{eq:respart}) can be extended to alternative structure of networks beyond ResNet, and extract more residuals as shown in Fig.(\ref{fig:rr}). In plain networks, the residual features can be computed by calculating the difference between high and low level features. The pooling layer also loses much details, so in extended residual structure, we upsample the feature after pooling layer, and compute residual features with the feature before the pooling layer. Then through the whole networks, all information can be used for segmentation.

Then residuals of target pyramid can be predicted with the $(- \bm{res})$ features using the appended predictors. Compared with other feature reconstruction methods, $(- \bm{res})$ has little information overlapping with the main part of $\bm{x}_{L}$.

\subsection{Loss Function for Training}\label{loss}
As directly decomposing the labels into residual pyramid will cause class conflict and lead to mismatch, we train the reconstructed predicted target and make the network learn the residuals indirectly.

Therefore, we reconstruct the predicted target from top to bottom at training phase and use the reconstructed targets of different levels to compute losses with the scaled labels:
\begin{gather}
l_{i} = criterion(\bm{target}_{i} , \, \bm{label}_{i}) \label{eq:loss1}\\
\bm{target}_{i} = \bm{target}_{1} + \sum_{k=2}^{i}\bm{tres}_{k}\quad(i\geq2). \label{eq:main2}
\end{gather}
$\bm{tres}_{i}$ and $\bm{target}_{i}$ are predicted residuals and reconstructed targets at the $i$-th level, $\bm{label}_{i}$ represents the scaled ground truth label of the $i$-th level using the nearest neighbor interpolation and $criterion()$ is Cross Entropy Loss. The upsampling of $\bm{target}_{i}$ is needed to add $\bm{target}_{i}$ with $\bm{tres}_{i+1}$, using bilinear interpolation.
In the same way, the residual pyramid of label can be defined as:
\begin{equation}
\bm{label}_{i} = \bm{label}_{1} + \sum_{k=2}^{i}\bm{lres}_{k}\quad(i\geq2), \label{eq:res2}
\end{equation}
where $\bm{lres}_{i}$ is the label residual of the $i$-th level. Then the learning process can be described as:
\begin{equation}
\min\; criterion(\bm{target}_{1} + \sum_{k=2}^{i}\bm{tres}_{k} ,\, \bm{label}_{1} + \sum_{k=2}^{i}\bm{lres}_{k}),\label{eq:loss5}
\end{equation}
Recursively, Eqn.(\ref{eq:loss5}) is equal to
\begin{equation}
\min\; criterion(\bm{tres}_{i} , \, \bm{lres}_{i}), \label{eq:lossres}
\end{equation}
which means that after being reconstructed to learn the scaled labels, the residual features are finally transformed into the label residuals.
The sum of residual pyramid losses is the final loss:
\begin{equation}
loss = \sum loss_{i},\label{eq:lossall}
\end{equation}
which indicates that we can train the losses seperately for different level of residuals.

There is multiple sub-networks of different depth in RPNet, and multiple sub-losses can help the propagation of gradient in backward pass, thereby facilitating the training process.

\renewcommand\arraystretch{1.25}
\setlength{\tabcolsep}{15pt}
\begin{table*}[t]
\begin{center}
\small
\begin{tabular}{l|c|c|c|c}
\hline
 & Name & Type & Output Channel & \\
\hline
\multirow{2}{*}{1/2 Scale} & Initial & & 16 &\multirow{2}{*}{Level-3\,(Res)}\\\cline{2-4}
& Bottleneck1.0 & Regular & 16 & \\
\hline
\multirow{5}{*}{1/4 Scale} & Bottleneck2.0 & Downsampling & 64 &\multirow{5}{*}{Level-2\,(Res)}\\\cline{2-4}
& Bottleneck2.1 & Regular & 64 & \\
& Bottleneck2.2 & Regular & 64 & \\
& Bottleneck2.3 & Regular & 64 & \\
& Bottleneck2.4 & Regular & 64 & \\
\hline
\multirow{11}{*}{1/8 Scale} & Bottleneck3.0 & Downsampling & 128 &\multirow{11}{*}{Level-1\,(main)}\\\cline{2-4}
& Bottleneck3.1 & Regular & 128 & \\
& Bottleneck3.2 & Regular & 128 & \\
& Bottleneck3.3 & Asynnetric & 128 & \\
& Bottleneck3.4 & Regular & 128 & \\
& Bottleneck3.5 & Regular & 128 & \\
& Bottleneck3.6 & Regular & 128 & \\
& Bottleneck3.7 & Asynnetric & 128 & \\
& Bottleneck3.8 & Regular & 128 & \\\cline{2-4}
& \multicolumn{3}{c|}{Repeat section3,without bottleneck3.0} & \\\cline{2-4}
& MainOut & $Conv$ & classes & \\
\hline
\end{tabular}
\end{center}
\vspace{-.5em}
\caption{Backbone network of the RPNet using reproduced ENet encoder~\cite{Paszke2016ENetAD}.}
\vspace{-.5em}
\label{tab:Backbone network}
\end{table*}

\subsection{The residual predictors}\label{section:rp}

The residual predictors followed by residual features are used to predict different levels of residuals. We design two kinds of predictors shown in Fig.(\ref{fig:rr}): basic type and guided type. Basic predictor simply use 1$\times$1 convolutions to predict main part and residuals. Guided type uses the main part feature of last-level to guide the prediction of residuals and uses several 1$\times$1 to adjust the channels. Both of the predictors follow the idea of Section.(\ref{loss}) to predict residuals indirectly.

As the traning process is step by step, when training one level of residuals, the main part feature has been trained to be able to recognize the pixel at higher level. We upsample the main feature and concatenate it with bigger residual features, then the combined feature will be easy to train, and lead to a better result. More details can be found in Section.(\ref{section:ab}).

\subsection{Process of Training a RPNet}
As residual target is the subtraction of the full target at this level and the main output of last level, to ensures that the residual feature focuses on the residual target, we should train the ResBlocks of different levels from top to down step by step. Actually, when we train residual features at last level, the main feature has been well trained to main target. If both of residual and main features are trained from scratch at same time, the targets for residual and main feature will be ambiguous and the network will not converge to optimal solution. The results of evaluation in Table.(\ref{tab:losscomp}) also proves validity of our training method. \\
The loss function for $i$-th level is\\
\begin{equation}
Loss_{k} = \sum_{i=1}^{k} loss_{i}.\nonumber
\end{equation}

\begin{figure}[hpt]
\begin{center}
   \includegraphics[width=1\linewidth]{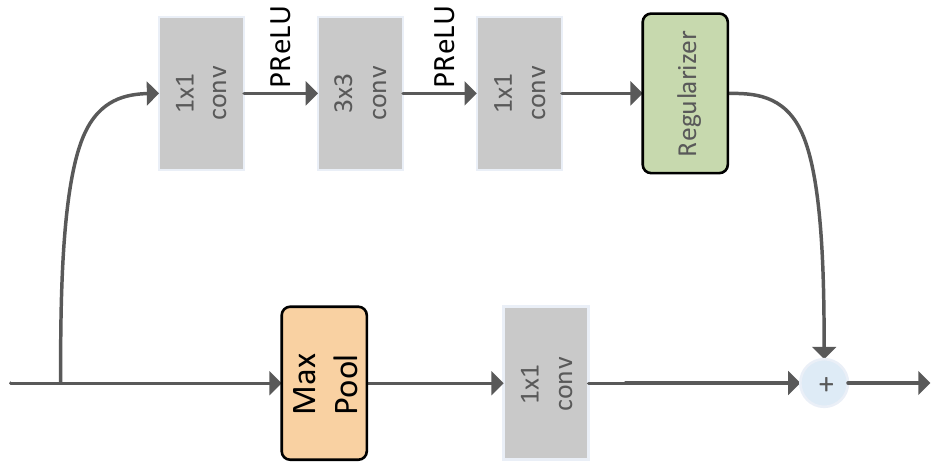}
\end{center}
   \caption{Downsampling block used in reproduced ENet~\cite{Paszke2016ENetAD}.}
\label{fig:downsample}
\end{figure}

Take the ENet + RPNet as an example, The backbone network is shown in Table.(\ref{tab:Backbone network}). The Initial and Bottleneck blocks are the same as that proposed in~\cite{Paszke2016ENetAD}, and we reproduce ENet with new Downsampling block shown in Fig.(\ref{fig:downsample}).

\begin{figure}[htp]
\begin{center}
   \includegraphics[width=1\linewidth]{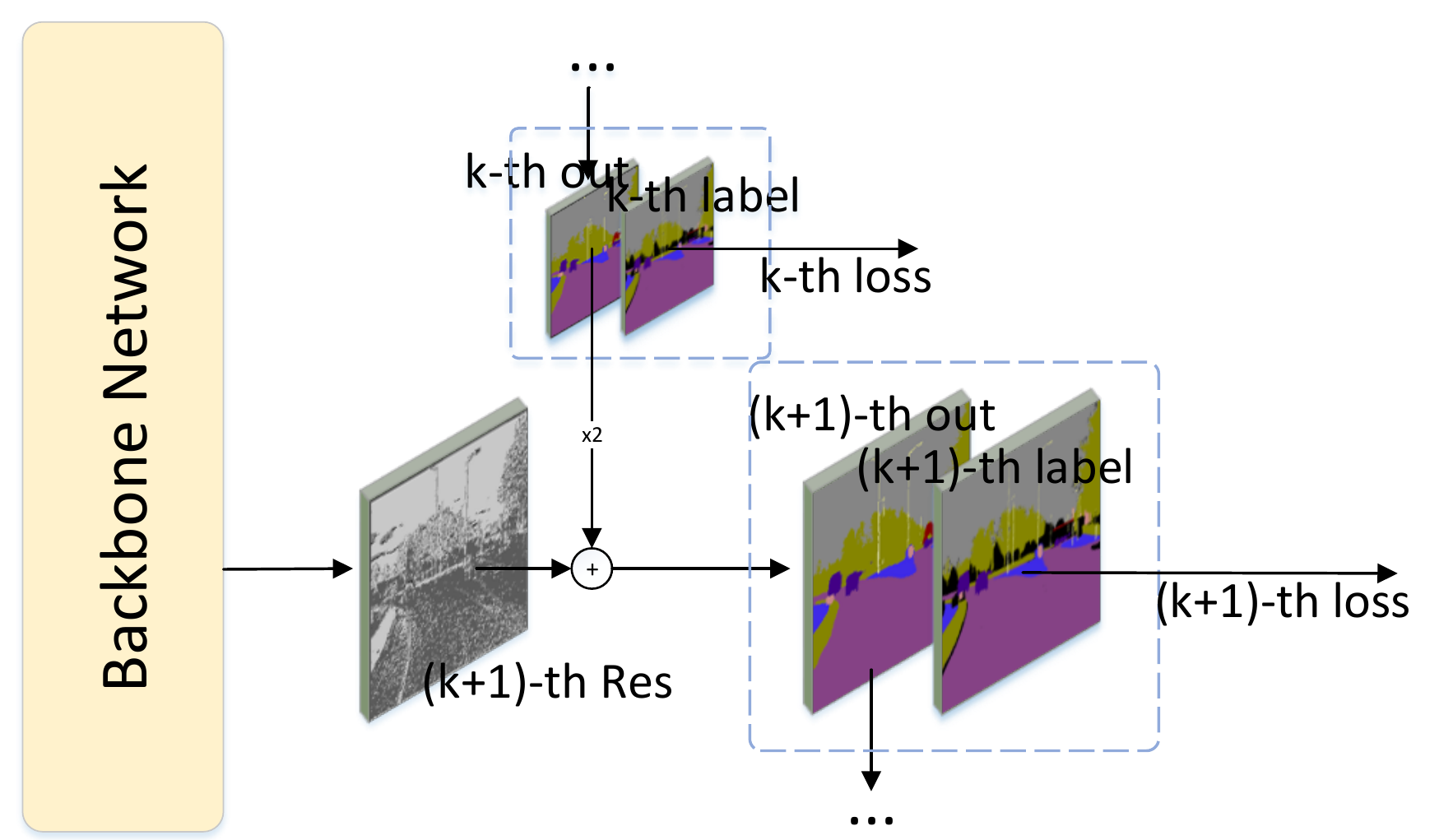}
\end{center}
   \caption{Top-down level-wise training of RPNet (basic predictor).}
\label{fig:loss}
\end{figure}

First, train the main prediction of 1/8 scale of original inputs; second, upsample the main prediction using bilinear interpolation; next execute the pixel-sum between the main; then use the Res2 to train the prediction of 1/4 scale of original inputs, the same way for 1/2 scale; finally, we get a residual pyramid and in order to save computing resources, we directly upsample the 1/2 scale predicted result to get full scale segmentation. The visualization of training process is shown in Fig.(\ref{fig:loss}).

\section{Experiments}
A set of experiments is conducted to verify the effectiveness and the efficiency. All experiments are evaluated on NVIDIA GTX 1080Ti and NVIDIA JETSON TX2. The main datasets in the experiments are from CamVid~\cite{Brostow2009SemanticOC} and Cityscapes~\cite{Cordts2016TheCD}.

\noindent\textbf{CamVid}
The CamVid is a street scene dataset from the perspective of driving automobile, which consists of 701 images with the resolution of 480$\times$360. The CamVid has 367 images for training, 233 for testing, including 11 classes.

\noindent\textbf{Cityscapes}
The Cityscapes is also a street scene dataset with 5000 fine-annotated images, 2975 for training, 500 for validation and 1525 for testing, at the resolution of 2048$\times$1024, including 19 classes. We only use the fine-annotated images for training, downsample the original images to 1024$\times$512 for training, and interpolate the outputs to 2048$\times$1024 for testing.

\subsection{Implementation}
\noindent\textbf{Network}

To show the performance of RPNet, we select ENet and ERFNet~\cite{Paszke2016ENetAD,Romera2018ERFNetER} as our baseline models. ENet is one of the fastest models on cityscapes benchmark, and ERFNet~\cite{Romera2018ERFNetER} is a little bit slower but more accurate. Both of the models have ResBlock in their encoder. The decoders of the two models are removed and the encoders are reformed as RPNet. We also set up a model of an encoder of ENet with a bilinear interpolation directly interpolate to original size as a comparative model. To construct the level-3 residual, we add a regular bottleneck1.0 to original ENet encoder.

\noindent\textbf{Training details}

We use Adam optimization with decay 0.0001 and batch size 3. We apply the poly learning rate policy, and the learning rate is multiplied by:
\begin{equation}
  (1-iter/max_iter)^{power}\nonumber
\end{equation}
 with power 0.9 and initial learning rate 0.0005. Class weighting scheme is:
 \begin{equation}
   \mathcal{W}_{class}=1/log(\mathcal{P}_{class}+k),\nonumber
 \end{equation}
 where k is set to 1.12.

\noindent\textbf{Data augmentation}

We use random horizontal flip and the transition of 0\~{}2 pixels on both axes of the input images to augment the dataset at training time.

\subsection{Ablation Study}\label{section:ab}

\newcommand{\resolution}[2]{\multicolumn{2}{c}{#1$\times$#2}}
\newcommand{\ms}{\multicolumn{1}{c}{ms}}
\newcommand{\fps}{\multicolumn{1}{c}{fps}}
\setlength{\tabcolsep}{8pt}
\begin{table*}[ht]
  \begin{center}
  \small
  \begin{tabular}{ l r r r r r r r r r r r r r r }
    \toprule
    \multirow{3}{*}{Model} &&\multicolumn{6}{c}{NVIDIA TX1} &  &\multicolumn{6}{c}{NVIDIA GTX 1080Ti} \\
    \cmidrule{3-8} \cmidrule{10-15}
            &&\resolution{480}{320} &\resolution{640}{360}&\resolution{1280}{720} &
            &\resolution{640}{360}&\resolution{1280}{720}&\resolution{1920}{1080} \\
            &&\ms &\fps &\ms &\fps &\ms &\fps & &\ms&\fps &\ms&\fps &\ms&\fps \\
    \midrule
    ENet  &&55    &18     &74   &13   &249    &4    & &8.8     &114   &10  &100    &26.2    &38     \\
    RPNet-sr-bp    &&47     &22    &60     &16  &200  &5  & &6.5      &154  &6.7   &149   &14.3     &70   \\
    \bottomrule
  \end{tabular}
  \end{center}
  \vspace{-.5em}
  \caption{Speed comparison of our method against the baseline of different input sizes on edge (TX2) and desktop (1080Ti) platforms.}
  \vspace{-.5em}
  \label{tab:speed}
\end{table*}

We reproduce ENet~\cite{Paszke2016ENetAD} and compare the different settings: full ENet, ENet encoder with interpolation and ENet encoder with RPNet. The RPNet is trained in a level-wise way. We evaluate the methods on the CamVid test dataset with PASCAL VOC intersection-over-union metric (IoU)~\cite{Everingham2009ThePV}. The result is shown in Table.(\ref{tab:ablition}).

\renewcommand\arraystretch{1.35}
\setlength{\tabcolsep}{4pt}
\begin{table}[htp]
\begin{center}
\small
\begin{tabular}{l|c|c|c}
\hline
 Method & FLOPs & Parameters & Mean IoU \\
\hline
 ENet & 3.00B & 0.3890M & 59.89 \\
\hline
 ENet encoder & 2.62B & 0.3507M & 59.60 \\
\hline
 RPNet-sr-bp(L2) & 2.64B & 0.3514M & 62.31 \\
 \hline
 RPNet-sr-bp(L2,L3) & 2.65B & 0.3516M & 63.29 \\
\hline
 RPNet-sr-gp & 2.73B & 0.3542M & 63.90 \\
\hline
 RPNet-er-bp & 2.65B & 0.3516M & 64.04 \\
 \hline
 RPNet-er-gp & 2.73B & 0.3542M & 64.67 \\
 \hline
\end{tabular}
\end{center}
\vspace{-.5em}
\caption{Speed and parameters analysis of the ENet, ENet encoder and ENet with RPNet at Level2(L2) and Level3(L3).}
\vspace{-.5em}
\label{tab:ablition}
\end{table}

The absence of the decoder structure helps the ENet encoder be faster than full ENet but at the same time results in a lower level of accuracy. In RPNet, although residual predictors are added, the FLOPs and parameters are almost at the same level as the ENet encoder. Such result is thanks to the reduction of depth and increase of width of the network, which improves the performance in device parallel. The RPNet adds the details to segmentation and makes a great improvement on accuracy, where the higher residuals has greater contributions to the results.

\begin{figure*}[hpt]
\begin{center}
   \includegraphics[width=0.8\linewidth]{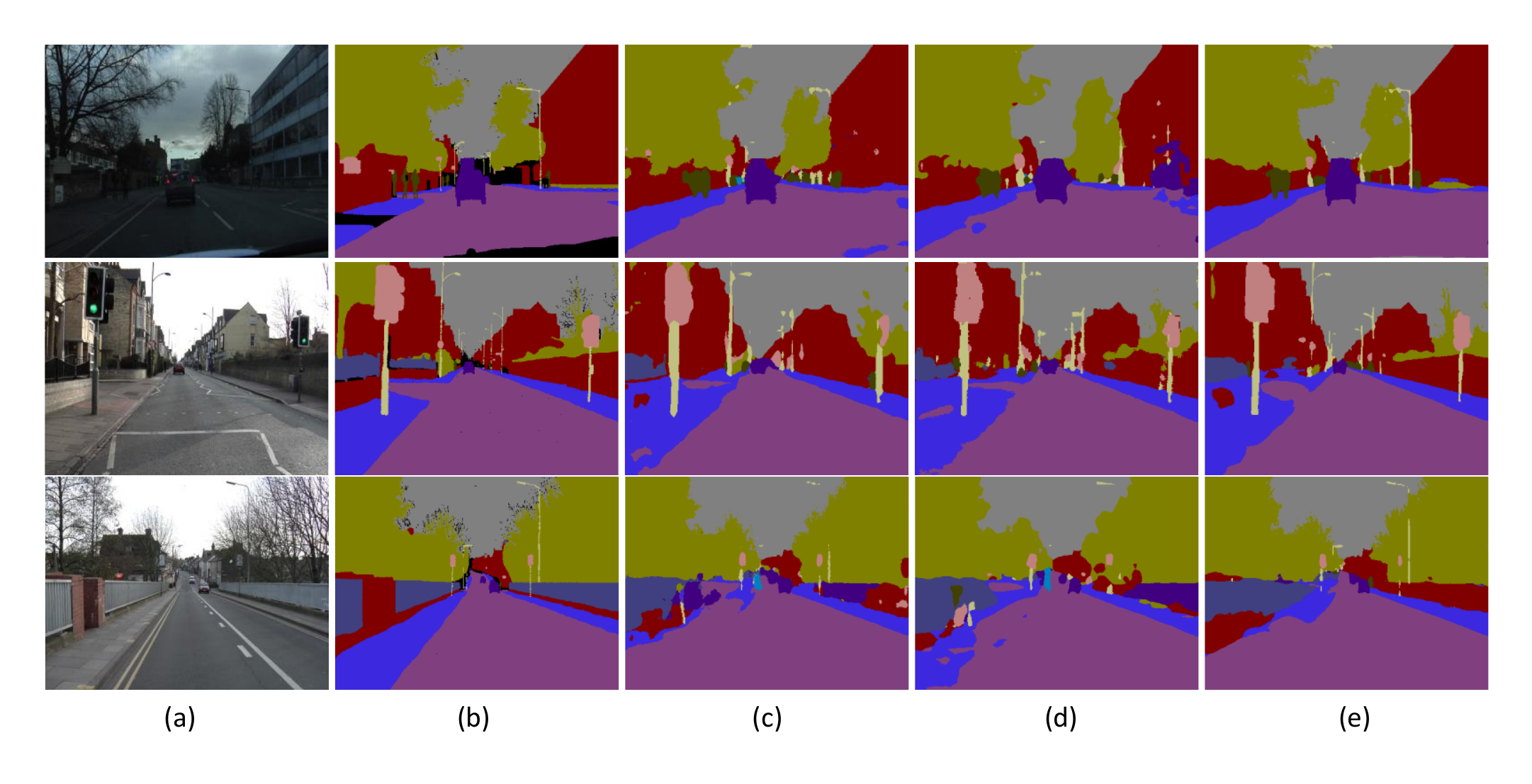}
\end{center}
   \caption{Sample results on the CamVid (480$\times$360) test dataset. From left to right: (a) Input, (b) Ground truth, (c) ENet, (d) ENet encoder, (e) ENet + RPNet.}
\label{fig:samples_cam}
\end{figure*}

From another aspect, decoder of original ENet has a limited increase of $0.29$ on mean IoU, while RPNet is at least 3.4 higher on ENet. The RPNet-er-gp even has a 4.78 improving on ENet. The RPNet has a distinct advantage on accuracy in the visualization of sample results in Fig.(\ref{fig:samples_cam}).

We also compare the different structures of residuals and predictors. Expanded residuals (\textbf{er}) extract more details from downsampling process. Though not increasing parameters or computation, it enhance the performance significantly. Base predictor (\textbf{bp}) of 1$\times$1 convolution is proved to be effective for residuals as the combination of sr and \textbf{bp} has shown a performance boost. When replaced the \textbf{bp} with guided predictor (\textbf{gp}), the RPNet will be further improved. Finally, the best combination of RPNet is proved to be \textbf{er} and \textbf{gp}.

In parameters and FLOPs comparison, all RPNet has advantages with original encoder-decoder structure. At the same time, the improvement on mean IoU is also significant.

Then to study on the effect on different training methods of the RPNet, we compare the single training with different type of weighted losses mentioned in Impatient DNNs~\cite{DBLP:journals/corr/AmthorRD16} and our level wise training on RPNet-sr-bp. The result is shown in Table.(\ref{tab:losscomp}).

\renewcommand\arraystretch{1.35}
\setlength{\tabcolsep}{5pt}
\begin{table}[hp]
\begin{center}
\small
\begin{tabular}{lccccc}
\hline
          & \textbf{EQ} & \textbf{LIN} & \textbf{POLY} & \textbf{NORM} & \textbf{Level-Wise} \\
\hline
\hline
 Mean IoU & 60.64 & 60.48 & 60.70 & 60.98 & 63.29 \\
\hline
\end{tabular}
\end{center}
\vspace{-.5em}
\caption{Comparison of single training with weighted losses and level-wise training.}
\vspace{-.5em}
\label{tab:losscomp}
\end{table}

In Table.(\ref{tab:losscomp}), the uniform weights(\textbf{EQ}), linearly increasing weights(\textbf{LIN}), polynomially increasing weights(\textbf{POLY}) and normal distribution weights \textbf{NORM}, described in~\cite{DBLP:journals/corr/AmthorRD16} are used to train RPNet. However, as analysed in Section(\ref{section:rp}), when we train the different levels from scratch together, though different type of weighted losses applied, the result of mean IoU will be worse than that with top-down level-wise training.

As shown in Fig.(\ref{fig:samples_cam}), the RPNet retains more details than original Enet structures, especially for small and thin objects such as signs and traffic lights, which is retrieved by the independent residual features of different levels.

We also compare the speeds of ENet and RPNet-sr-bp on different platforms, as shown in Table.(\ref{tab:speed}), which indicates that RPNet is more efficient when the computing resources are limited or the input size is large. On TX2, RPNet has an average $23\%$ promotion on fps, and on 1080Ti, RPNet has a $35\%
\sim 84\%$ increase of the resolution from $640\times360$ to $1920\times1080$.

As the computation from network depth is reduced and extra computation from network width is cheap, RPNet structure makes better use of computing resources. Note that the implementation on TX2 does not contain any extra acceleration tool such as TensorRT, the further-improved RPNet can be deployed in embedded real-time systems.

\begin{figure*}[hbt]
\begin{center}
   \includegraphics[width=1\linewidth]{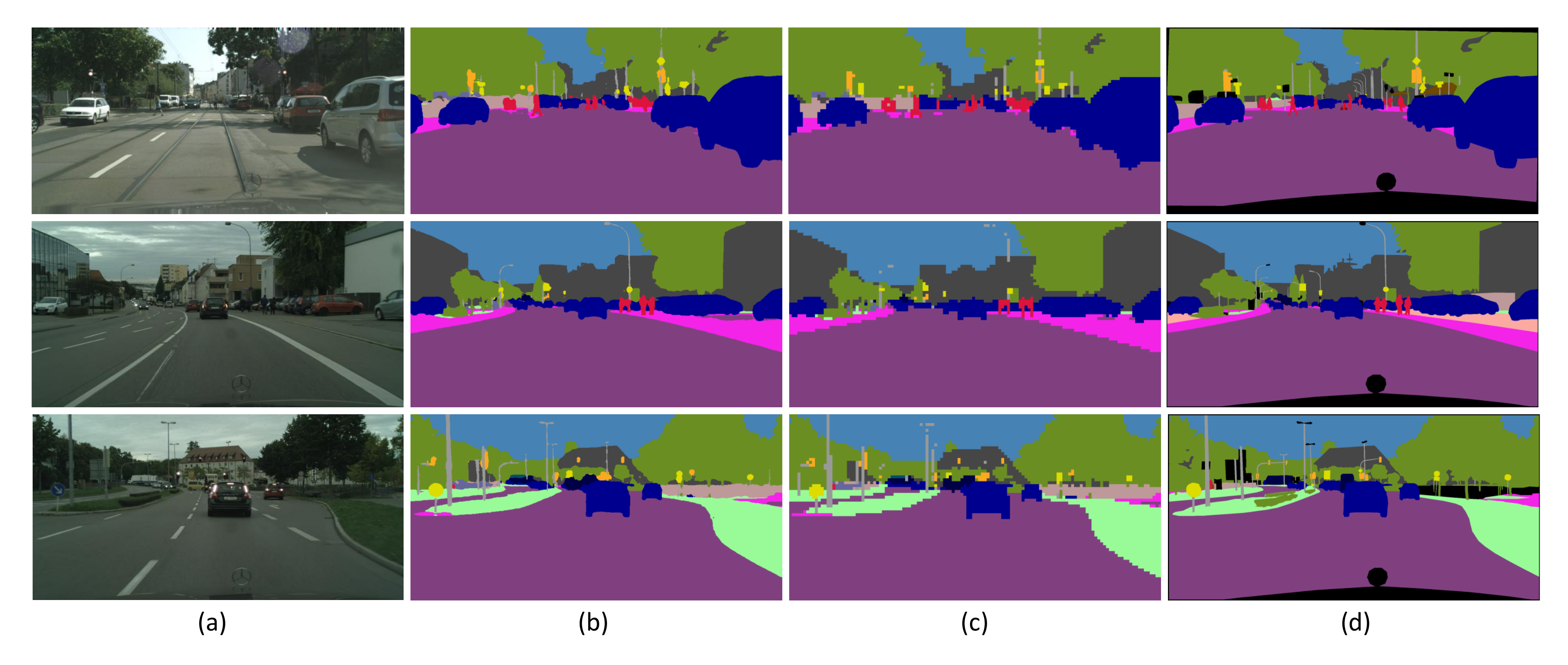}
\end{center}
   \caption{Sample results on the Cityscapes validation dataset. From left to right: (a) Input, (b) Level-3 output, (c) Level-1 output, (d) Ground truth.}
\label{fig:samples}
\end{figure*}

\subsection{Evaluation on CamVid dataset}
PASCAL VOC intersection-over-union metric(IoU)~\cite{Everingham2009ThePV} is used to evaluate the methods on CamVid and Cityscapes. An extra sematic instant-level intersection-over-union metric(iIoU) is used on Cityscapes, which focuses on how well the individual instances in the scene are represented in the labeling. In this section, we use expanded residuals and guided predictor as default setting of RPNet.

We construct RPNet on ENet and ERFNet encoders to evaluate the performance. ENet has 87 $Conv$ layers and ERFNet has 75 $Conv$ layers in max depth. After adding RPNet, the max depth comes to 71 and 55 and the parameters and FLOPs comparison are shown in Table.(\ref{tab:CamVid}).

\renewcommand\arraystretch{1.05}
\setlength{\tabcolsep}{5pt}
\begin{table}[hp]
\begin{center}
\small
\begin{tabular}{lcccc}
\hline
 Method & Mean IoU & fps & Parameters & FLOPs \\
\hline
 ENet~\cite{Paszke2016ENetAD} & 59.89 & 111 & 0.39M & 1.50B \\
 ERFNet~\cite{Romera2018ERFNetER} & 60.54 & 133 & 2.07M & 8.43B \\
 ESPNet~\cite{DBLP:journals/corr/abs-1803-06815} & 62.6 & 205 & 0.68M & 0.87B \\
 FC-DenseNet56~\cite{8014890} & 58.9 & 27 & 1.5M & 26.29B \\
 \hline
 \hline
 RPNet(ENet) & 64.67 & 102 & 0.35M & 1.36B \\
 RPNet(ERFNet) & 64.82 & 149 & 1.89M & 6.78B \\
\hline
\end{tabular}
\end{center}
\vspace{-.5em}
\caption{Performance and computation comparison on CamVid (480$\times$360).}
\vspace{-.5em}
\label{tab:CamVid}
\end{table}

\renewcommand\arraystretch{1.35}
\setlength{\tabcolsep}{10pt}
\begin{table*}[htp]
\begin{center}
\small
\begin{tabular}{lccccc}
\hline
 Method & Input Size & Mean IoU & Mean iIoU & fps & FLOPs \\
\hline
 ENet~\cite{Paszke2016ENetAD} & 1024$\times$512 & 58.3 & 34.4 & 77 & 4.03B \\
 ERFNet~\cite{Romera2018ERFNetER} & 1024$\times$512 & 68.0 & 40.4 & 59 & 25.60B \\
 ESPNet~\cite{DBLP:journals/corr/abs-1803-06815} & 1024$\times$512 & 60.3 & 31.8 & 139 & 3.19B \\
 BiSeNet~\cite{Yu2018BiSeNetBS} & 1536$\times$768 & 68.4 & - & 69 & 26.37B \\
 ICNet~\cite{zhao2018icnet} & 2048$\times$1024 & 69.5 & - & 30 & - \\
 DeepLab(MobileNet)~\cite{Chen2018EncoderDecoderWA} & 2048$\times$1024 & 70.71 (val) & - & 16 & 21.27B \\
 LRR~\cite{Ghiasi2016LaplacianRA} & 2048$\times$1024 & 69.7 & 48.0 & 2 & - \\
 RefinNet~\cite{Lin:2017:RefineNet} & 2048$\times$1024 & 73.6 & 47.2 & - & 263B \\
 \hline
 \hline
 RPNet(ENet) &1024$\times$512& 63.37 & 39.0 & 88 & 4.28B \\
 RPNet(ERFNet) &1024$\times$512& 67.9 & 44.9 & 123 & 20.71B \\
\hline
\end{tabular}
\end{center}
\vspace{-.5em}
\caption{Speed and accuracy comparison on Cityscapes.}
\vspace{-.5em}
\label{tab:Cityscapes}
\end{table*}

The Table.(\ref{tab:CamVid}) indicates that on small inputs with adequate computing resources, ERFNet, which has fewer layers in max depth but more FLOPS, is faster than ENet. But on larger inputs of Cityscapes, shown in Table.(\ref{tab:Cityscapes}), ERFNet with more FLOPS is slower than ENet. After replacing the decoder with RPNet, the FPS increases greatly, especially for RPNet(ERFNet) whose speed even exceeds the ENet. Besides, compared with decoder version of methods, the RPNet also has advantage on mean IoU.

\subsection{Evaluation on Cityscapes dataset}
The proposed RPNet achieves impressive results compared with the state-of-the-art methods. The iIoU of RPNet increases significantly compared with the baselines, which proves that residual features exactly enhance the details and small objects with residual prediction once again. The improved architecture of segmentation network also increases efficiency.

Besides, Fig.(\ref{fig:samples}) shows the intermediate and final results of RPNet. We can find the lost details retrieved by the RPNet by comparing the column (b) and (c).

In larger inputs, heavy networks show the advantage in learning more characteristics, and the improvement of performance is greater on cityscapes compared with thin networks. ENet, ESPNet and RPNet (ENet) with FLOPs under 9B have lower IoUs than that with higher FLOPs, and RPNet still delivers a better performance. BiSeNet with with two-path structure also has the advantage on both accuracy and efficiency as good as ERFNet. The high speed of BiSeNet also comes from the shallower depth of network whose backbone network is lightweight Xception39~\cite{chollet2017xception}. Still, the bigger inference size of inputs limits the faster speed. ICNet has multi-resolution paths also has high mean IoU, but multiple paths also produce more computation, which makes the ICNet slower.

 LRR is much slower because of the heavy backbone networks and two-part prediction at testing time, despite of the 1.8 higher than RPNet (ERFNet) of IoU. RefineNet also achieves high accuracy but has much computation in high-resolution feature reconstruction. However, both LRR and RefineNet have superiority on iIoU as both of them use low-level feature to refine boundary and details of the targets. DeepLab (MobileNet) and ESPNet have less FLOPs but are slower than the same level methods. The reduction of FLOPs is attributed to the depth-wise separate convolution, but such operators are unable to make full use of computing resources.

\section{Conclusion}
In this paper, we have proposed a single-shot method for semantic segmentation based on Residual Pyramid Network (RPNet), which constructs ResBlock to learn residuals of different levels of the target. On this basis, we introduce variations of residual structure and predictor. With the residual learning blocks, the RPNet has better performance compared with methods with complicated feature reconstruction and well-designed decoder structures. At the same time, the single-shot structure makes the RPNet as fast as methods without decoder. Compared with the conventional structure models, RPNet delivers better performance on both efficiency and accuracy. The proposed RPNet is suitable to improve segmentation models with arbitrary structure networks which delete decoder structure and implement single-shot segmentation.

In our experiments, the RPNet is trained step-by-step. Though the test efficiency is improved, but the train period is long. The future works will involve experiments about the learning policy for RPNet to improve efficiency of train phase.


%

\section*{Acknowledgment}

This work was supported by the Natural Science Foundations of China (61727802 and BE2018126).

\ifCLASSOPTIONcaptionsoff
  \newpage
\fi



%

{\small
\bibliographystyle{ieee}
\bibliography{egbib}
}

%

\end{document}